\documentclass[letterpaper]{article} 
\usepackage{aaai24}  
\usepackage{times}  
\usepackage{helvet}  
\usepackage{courier}  
\usepackage[hyphens]{url}  
\usepackage{graphicx} 
\usepackage{natbib}  
\usepackage{caption} 
\usepackage{marvosym}
\usepackage{algorithm}
\usepackage{algorithmic}

\usepackage{newfloat}
\usepackage{listings}
\DeclareCaptionStyle{ruled}{labelfont=normalfont,labelsep=colon,strut=off} 
\floatstyle{ruled}
\newfloat{listing}{tb}{lst}{}
\floatname{listing}{Listing}
\usepackage{booktabs}
\usepackage{multirow}
\usepackage{nccmath}
\usepackage{amssymb}
\usepackage{cite}
\usepackage{graphicx}
\usepackage{amsmath}
\title{Brush Your Text: Synthesize Any Scene Text on Images via Diffusion Model}
\author{
    Lingjun Zhang\textsuperscript{\rm 1,\rm 2}\thanks{Equal Contribution.}\thanks{Work done as an intern at Shanghai AI Laboratory.},
    Xinyuan Chen\textsuperscript{\rm 2}\footnotemark[1],
    Yaohui Wang\textsuperscript{\rm 2},
    Yue Lu\textsuperscript{\rm 1}\thanks{Corresponding author.},
    Yu Qiao\textsuperscript{\rm 2}
}

\affiliations{
    \textsuperscript{\rm 1}East China Normal University, Shanghai, China\\
    \textsuperscript{\rm 2}Shanghai Artificial Intelligence Laboratory, Shanghai, China\\
    {\small \url{https://github.com/ecnuljzhang/brush-your-text}}

}

\usepackage{bibentry}

\begin{document}
\maketitle
\begin{figure*}[!ht]
    \centering
    \includegraphics[width=0.95\textwidth]{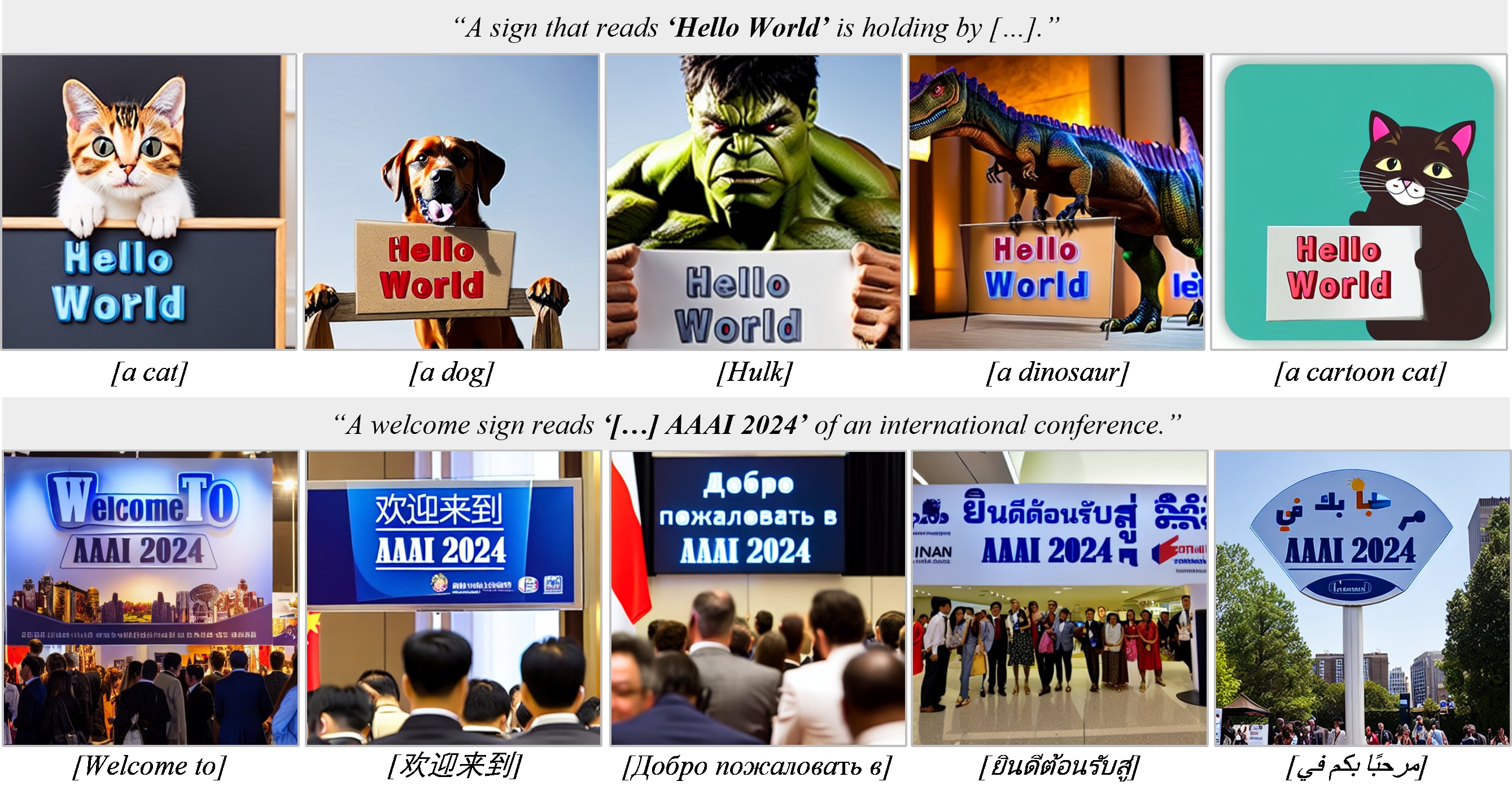}
    \caption{Diff-Text has the ability to generate accurate and realistic scene text images from a given scene text of any language along with a textual description of any scene.}
    \label{fig:teaser}
\end{figure*}

\begin{abstract}
Recently, diffusion-based image generation methods are credited for their remarkable text-to-image generation capabilities, while still facing challenges in accurately generating multilingual scene text images. To tackle this problem, we propose \textbf{Diff-Text}, which is a \textit{training-free} scene text generation framework for any language. Our model outputs a photo-realistic image given a text of any language along with a textual description of a scene. The model leverages rendered sketch images as priors, thus arousing the potential multilingual-generation ability of the pre-trained Stable Diffusion. Based on the observation from the influence of the cross-attention map on object placement in generated images, we propose a localized attention constraint into the cross-attention layer to address the unreasonable positioning problem of scene text. Additionally, we introduce contrastive image-level prompts to further refine the position of the textual region and achieve more accurate scene text generation. Experiments demonstrate that our method outperforms the existing method in both the accuracy of text recognition and the naturalness of foreground-background blending.
\end{abstract}

\section{Introduction}
Minority languages, such as Arabic, Thai, and Kazakh, not only have a significant number (reaching 5000 to 7000), but their low-resource nature also impedes the progress of computer vision, particularly in the domain of image generation. In recent years, with the advancement of diffusion models \cite{ho2020denoising}, significant progress has been made in generating realistic and prompt-aligned images~\cite{rombach2022high,ramesh2022hierarchical,saharia2022photorealistic}. However, achieving accurate scene text generation remains challenging due to the fine-grained structure within the scene text. 

Recent efforts utilize diffusion models to overcome the limitations of traditional methods and enhance text rendering quality. For instance, Imagen \cite{saharia2022photorealistic} and DeepFloyd \cite{deepfloyd} use the T5 series to generate text better. While these methods are capable of generating structurally accurate scene text, they demand a large amount of training data which is not suitable for minority languages and still lack control over the generated scene text. Some researchers \cite{wu2019editing,yang2020swaptext,lee2021rewritenet,krishnan2023textstylebrush} exploit GAN \cite{goodfellow2014generative} based scene text editing methods to generate scene text, which is more controllable. However, these methods are confined to generating scene text at the string level and do not possess the capability to generate complete scene compositions.

To tackle these challenges, we propose a training-free framework, Diff-Text, and a simple yet highly effective approach for multilingual scene text image generation. Our proposed framework inherits the off-the-shelf diffusion model while specializing in text generation by localized attention constraint method along with positive and negative image-level prompts. Specifically, given a text to be rendered, we first render it to a sketch image and then detect the edge map which is used as the control input of our model. Our model generates a realistic scene image according to the control input and the prompt input which contains a description of a scene. However, the control inputs are easily treated as grotesque patterns instead of texts on signs or billboards. Recent research \cite{hertz2022prompt} suggests that the input prompts exert their influence on the object placement within the generated images via the cross-attention mechanism. Inspired by this observation, we first identify the keywords in the prompt that correspond to the textual region, such as ``sign", ``notice", and ``billboard", and then constrain the cross-attention maps for these keywords to the textual region. Furthermore, we introduce a positive image-level prompt that further refines the placement of the textual region and a negative image-level prompt that enhances the alignment between the generated scene text and edge image, thereby ensuring greater accuracy in the generated scene text. Experiments demonstrate the effectiveness and robustness of our method.

\section{Related Works}
\textbf{Scene Text Generation} automates the creation of scene text images from provided textual content. Notably, SynthText \cite{gupta2016synthetic} is widely used to train scene text recognition models. It employs existing models to analyze images, identifies compatible text regions in semantically coherent areas, and places processed text using a designated font. Furthermore, SynthText3D \cite{liao2020synthtext3d} and UnrealText \cite{long2020unrealtext} generate scene text images from a virtual realm using a 3D graphics engine. However, these methods directly overlay text onto the background, resulting in artifacts in text appearing, which leads to a significant disparity between the synthesized and real image distributions. Some methods introduce GANs for realistic image generation. SF-GAN \cite{zhan2019spatial} introduces geometry and appearance synthesizers for realistic scene text generation, but struggles with accurate text placement. Scene text editing methods \cite{wu2019editing,yang2020swaptext,roy2020stefann,zhang2021scene,lee2021rewritenet,Xie_2021_CVPR,krishnan2023textstylebrush,he2022diff} attempt tackle this problem. However, these methods concentrate only on generating the text region rather than the entire image.

\begin{figure*}[!htb]
    \centering
    \includegraphics[width=0.95\linewidth]{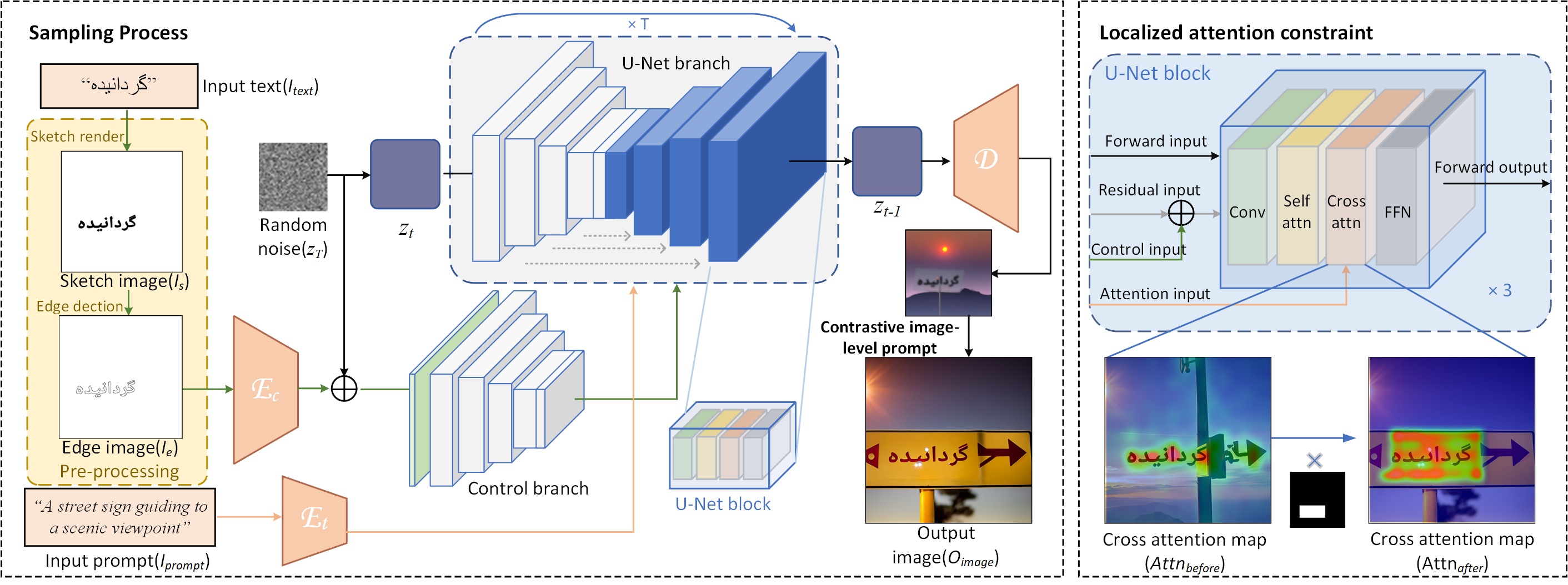}
    \caption{Our model employs input text ($I_{text}$) of any language to serve as the foreground element. The text is subsequently rendered into a sketch image, and its edges are detected to derive an edge image, which acts as an input of the control branch. Concurrently, our model takes in an input prompt ($I_{prompt}$) as the description of the background scene. After $T$ denoising iterations, the model generates the final output image ($O_{image}$). Localized attention constraint and contrastive image-level prompts are employed in the U-Net block's cross-attention layer to enhance textual region positioning for precise scene text generation.
    }
    \label{fig:framework}
\end{figure*}
\begin{figure}[h]
    \centering
\includegraphics[width=0.85\linewidth]{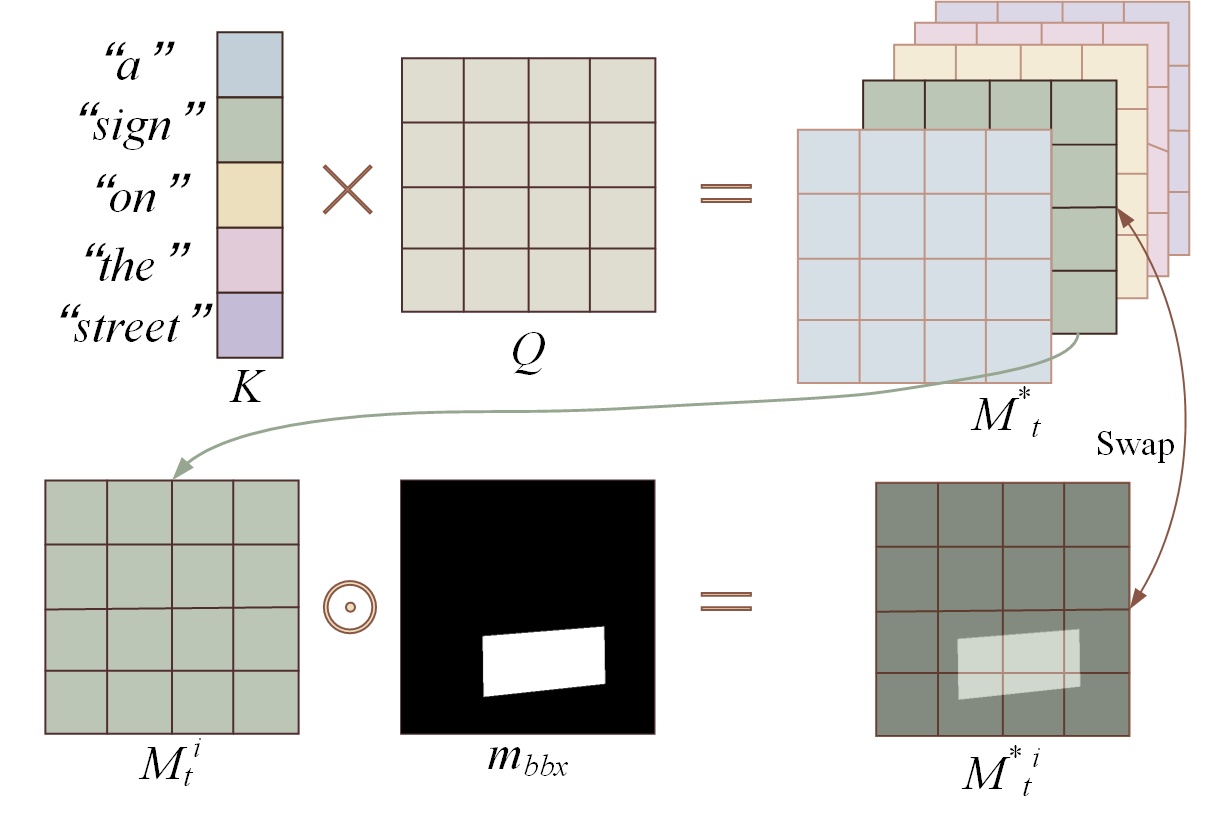}
    \caption{Details of the proposed localized attention constraint method. The ``$\times$" signifies matrix multiplication, while ``$\odot$" denotes element-wise multiplication.}
    \label{fig:attention_constraint}
\end{figure}

\textbf{Text-to-Image Generation} represents a promising result that has seen significant progress in generating realistic and prompt-aligned images \cite{rombach2022high,ramesh2022hierarchical,saharia2022photorealistic}, as well as videos \cite{makeavideo,imagenvideo,videoldm,pyoco,lavie,seine,leo}, through the application of diffusion models \cite{ho2020denoising}.
GLIDE \cite{nichol2021glide} introduces text conditions into the diffusion process using classifier-free guidance. DALL-E 2 \cite{ramesh2022hierarchical} adopts a diffusion prior module on CLIP text latent and cascaded diffusion decoder to generate high-resolution images. Imagen \cite{saharia2022photorealistic} emphasizes language understanding and proposes to use a large T5 language model for better semantics representation. Stable Diffusion \cite{rombach2022high} is an open-sourced model that projects the image into latent space with VAE and applies the diffusion process to generate feature maps in the latent level. 

In addition to text conditions, a realm of research explores controlling diffusion models through image-level conditions. Certain image editing methods \cite{meng2021sdedit,kawar2023imagic,mokady2023null,brooks2023instructpix2pix} introduce images to be edited as conditions in the denoising process. Image inpainting \cite{balaji2022ediffi,avrahami2022blended,lugmayr2022repaint,bau2021paint} constitutes another type of editing method, aiming to generate coherent missing portions of an image based on a specified region while preserving the remaining areas. Additionally, SDG \cite{liu2023more} represents an alternative approach involving extra conditions, which injects semantic input using a guidance function to direct the sampling process of unconditional DDPM. Some methods \cite{chen2023textdiffuser,ma2023glyphdraw} utilize textual layouts or masks as conditions for scene text generation. However, these approaches need extensive labeled datasets of scene text for training, which poses a challenge for low-resource languages.

Moreover, ControlNet \cite{zhang2023adding} and T2I-adapter \cite{mou2023t2i} are dedicated to offering a comprehensive solution for controlling the generation process by leveraging auxiliary information like edge maps, color maps, segmentation maps, \textit{etc}. These methods exhibit remarkable control and yield impressive results in terms of image quality. In this work, we perceive scene text generation as a text-to-image task with supplementary control (scene text) and incorporate the rendered scene text as an image-level condition within the diffusion model.
\begin{figure*}[!thb]
    \centering \includegraphics[width=0.92\textwidth]{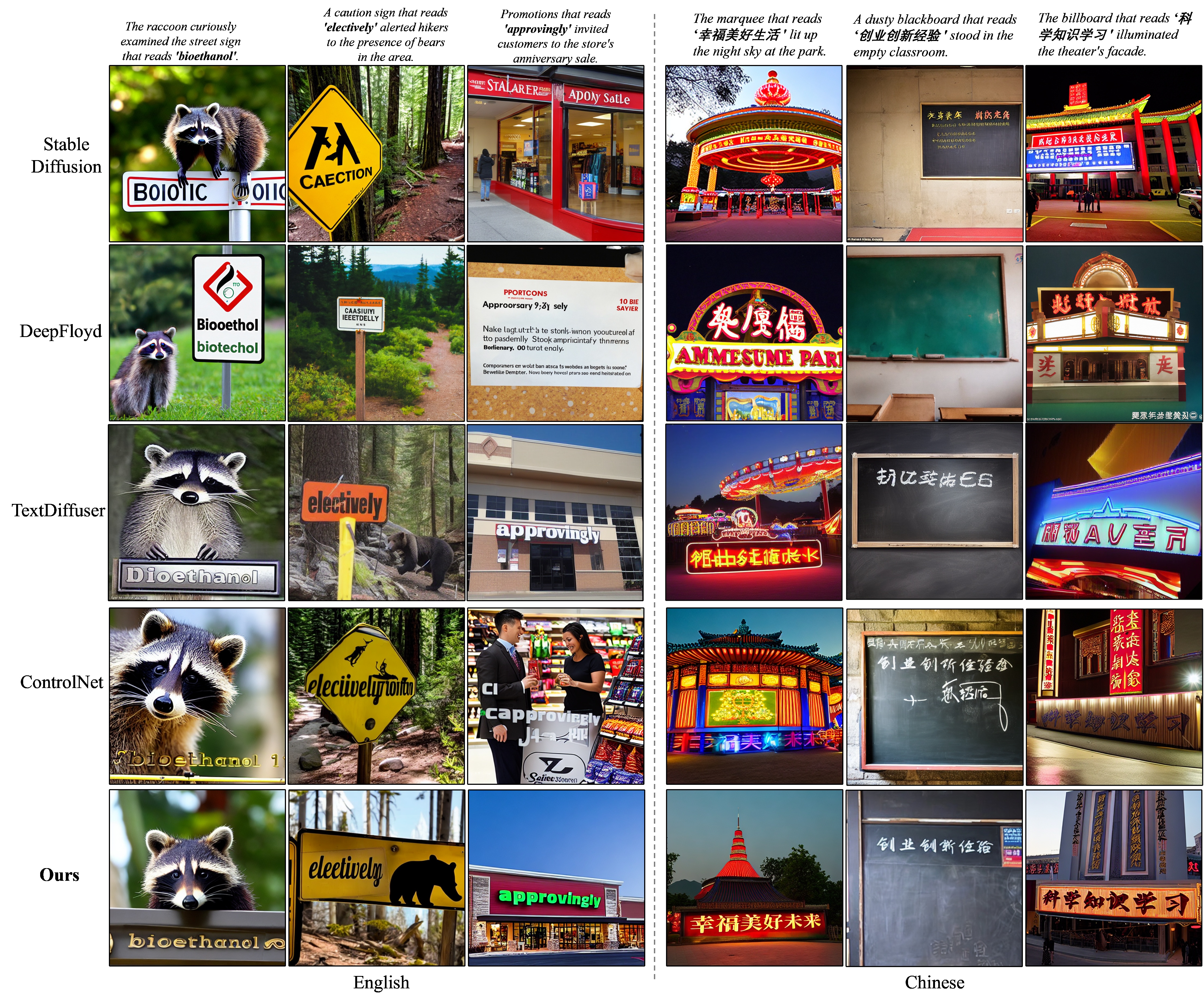}
    \caption{Visualizations of scene text generation in English and Chinese, compared with existing methods. The first three columns represent the generated results of English scene text, while the last three columns depict the generated results of Chinese scene text.} \label{fig:big_compare}
\end{figure*}
\begin{figure*}[!thb]
    \centering   \includegraphics[width=0.9\textwidth]{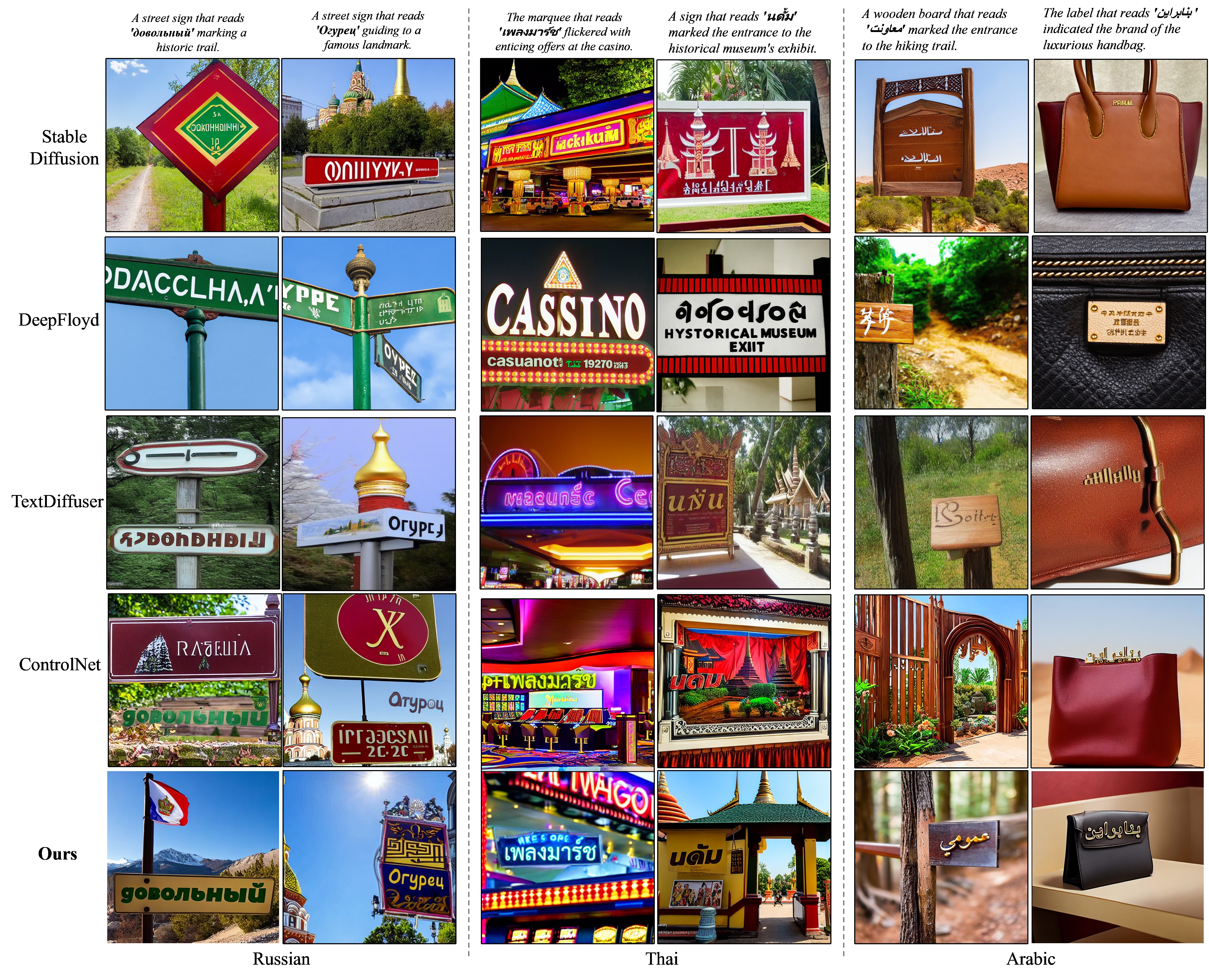}    \caption{Visualizations of scene text generation in Russian, Thai, and Arabic, compared with existing methods. The first and second columns present the results of Russian scene text, the third and fourth columns depict Thai scene text, and the final two columns illustrate Arabic scene text.}
    \label{fig:small_compare}
\end{figure*}
\begin{table*}
\centering
\setlength{\tabcolsep}{4.5mm}{
\scalebox{0.9}{
\begin{tabular}{ccccccc}
\toprule
Language                 & Metrics        & Stable Diffusion & DeepFloyd       & TextDiffuser    & ControNet & Ours            \\ \midrule
\multirow{3}{*}{\rotatebox{90}{Arabic}}  & CLIPScore      & 0.7961          & 0.7335          & 0.8084          & 0.8067    & \textbf{0.8138} \\
                         & Accuracy       & 0.000           & 0.000           & 0.000           & 3.291     & \textbf{33.13} \\
                         & Edit\_accuracy & 16.65           & 13.17           & 11.58           & 34.80     & \textbf{72.93}  \\ \midrule
\multirow{3}{*}{\rotatebox{90}{Thai}}    & CLIPScore      & 0.7733          & 0.7926          & 0.7873          & 0.8059    & \textbf{0.8164} \\
                         & Accuracy       & 0.000           & 0.000           & 0.000           & 7.160     & \textbf{38.41}  \\
                         & Edit\_accuracy & 10.70           & 14.51           & 11.64           & 36.34     & \textbf{82.97}  \\ \midrule
\multirow{3}{*}{\rotatebox{90}{Russian}} & CLIPScore      & 0.7948          & 0.8201          & 0.8335          & 0.8306    & \textbf{0.8632} \\
                         & Accuracy       & 0.000           & 0.000           & 1.375           & 9.790     & \textbf{39.29}  \\
                         & Edit\_accuracy & 14.60           & 26.05           & 37.72           & 39.21     & \textbf{80.58}  \\ \midrule
\multirow{3}{*}{\rotatebox{90}{English}} & CLIPScore      & 0.7879          & 0.8658          & \textbf{0.8666} & 0.7334    & 0.8649          \\
                         & Accuracy       & 0.083           & 16.67           & 43.91           & 12.88    & \textbf{61.03}  \\
                         & Edit\_accuracy & 32.75           & 66.20           & 84.84           & 40.04     & \textbf{89.52}  \\ \midrule
\multirow{3}{*}{\rotatebox{90}{Chinese}} & CLIPScore      & 0.8265          & 0.8347            & 0.8201     & 0.8312    & \textbf{0.8351}    \\
                         & Accuracy       & 0.000           & 0.000           & 0.000           & 5.875     & \textbf{32.40}   \\
                         & Edit\_accuracy & 3.890           & 6.830           & 9.598           & 26.81     & \textbf{68.75}  \\ \bottomrule
\end{tabular}}}
\caption{Quantitative comparison with existing methods across five languages. 
The bold numbers represent the best results among all compared methods.}
\label{table:quantitative}
\end{table*}
\section{Methods}
\subsection{Overall Framework}

We introduce a training-free scene text generation framework named Diff-Text, applicable to any language. Given an input text $I_{text}$ and a prompt $I_{prompt}$, our proposed framework can generate scene text images that encompass: (1) precise textual content of $I_{text}$; (2) scenes that align with the provided prompt $I_{prompt}$; and (3) seamless integration of textual content with the depicted scenes. The architecture of our framework is presented in Fig. \ref{fig:framework} and contains a pre-processing module, a U-Net branch, and a control branch.

Initially, the provided input text $I_{text}$ undergoes pre-processing and is rendered into a sketch image denoted as $I_s$, depicting black text against a whiteboard backdrop with a randomly chosen font. Subsequently, the Canny edge detection algorithm is applied to derive an edge image denoted as $I_e$. This image, serving as an image-level condition, is then utilized as input for the control branch. Simultaneously, the provided input prompt $I_{prompt}$ is processed by the text encoder, serving as a text-level condition. Under the guidance of both image-level and text-level conditions, the U-Net branch predicts the noise $z_t$ at time t and utilizes $z_t$ to reconstruct the output image from Gaussian noise.

Due to the independence of control input and prompt input for the U-Net network, there is a risk of incorrect fusion between image-level and text-level controls. For instance, the network might mistake the edges of the character ``O" as part of a circular pattern. This issue is particularly prominent in the generation of scene text images for minor languages. To address this concern, we introduce a localized attention constraint method tailored for scene text generation. Simultaneously, to ensure a more rational fusion and enhance the precision of image-level control, we have proposed a contrastive image-level prompt. The localized attention constraint is utilized to confine the cross-attention maps associated with text region descriptors from the prompt input, such as ``sign" or ``billboard". These maps are limited to areas near the text through a pre-processing module that generates random bounding boxes. Regarding the contrastive image-level prompt, it comprises a positive image-level prompt and a negative image-level prompt.

\begin{figure}[!ht]
    \centering
    \includegraphics[width=0.98\linewidth]{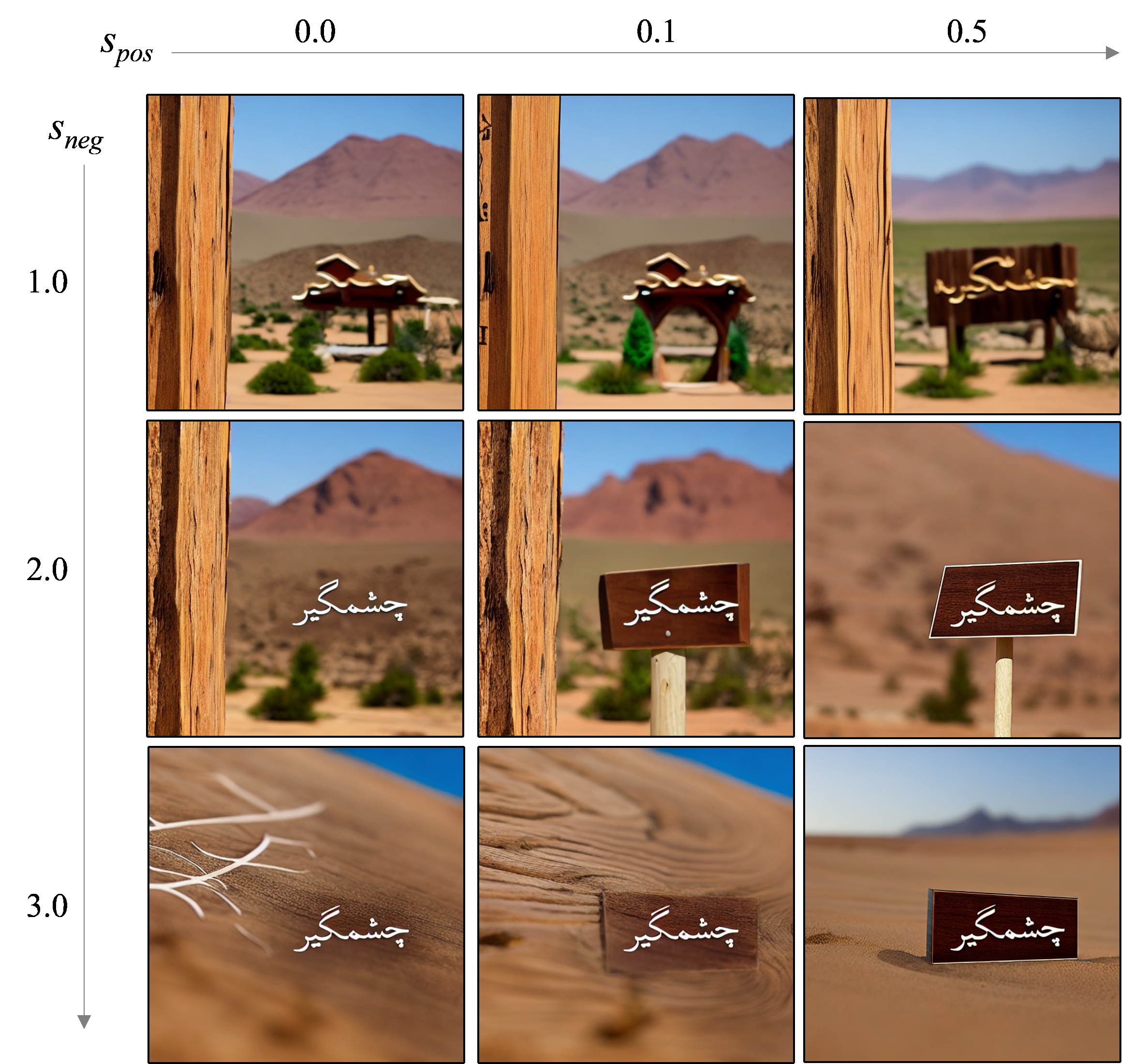}
    \caption{The image-level prompt comprises both positive and negative components, denoted as $s_{pos}$ and $s_{neg}$. $s_{pos}$ controls the intensity of ``sign" occurrences in the background, while $s_{neg}$ controls the clarity of the scene text.}
    \label{fig:IP_ablation}
\end{figure}
\subsection{Localized Attention Constraint}
Our goal is to place scene text sensibly within scenes, such as on billboards or street signs. To achieve this, we introduce the localized attention constraint method. 
As shown in Fig. \ref{fig:attention_constraint}, during one forward pass at each timestep, we traverse through all layers of the diffusion model and manipulate the cross-attention map. The cross-attention map is denoted as $M_t \in R^{HW \times d_t}$, where $HW$ refers to the width and height of $z_t$ at different scales, and $d_t$ represents the maximum length of tokens. In the framework, the positions of text within $I_{s}$ are either user-specified or randomly placed, which means, obtaining the corresponding text bounding box is straightforward. We use this bounding box to derive a mask image of the text region, which we define as $m_{bbx} \in R^{H \times W}$. Then, assuming that the indices of tokens corresponding to words that may contain text in the prompt are represented by the set $I$, we resize $m_{bbx}$ to $HW$ and compute the new cross-attention map $M^*_t = \{ \lambda \times M^i_t \odot m_{bbx} | \forall i \in I \}$. Finally, $M^*_t$ is involved in the calculation of the $z^*_{t-1}$. After applying the localized attention constraint, we find a sensible and appropriate position to place the scene text. This approach also enhances the natural integration of foreground text with the background, resulting in a more realistic scene text generation.

\subsection{Contrastive Image-level Prompts}
The limited availability of images for minority languages within the training dataset of Stable Diffusion frequently results in the misinterpretation of edge images as object outlines. This misinterpretation often leads to the introduction of additional strokes, ultimately resulting in unrecognizable scene text generation. 
Indeed, the effectiveness of the localized attention constraint method depends on the presence of objects in the generated image that can accommodate the placement of text. In other words, if $M^i_t, i \in I$ approaches 0 and $M^*_t$ remains the same as $M_t$, the localized attention constraint will not yield the desired output.

To tackle this issue, we introduce the definition of contrastive image-level prompt. In this regard, we consider the edge image $I_e$ as the foundation of the image-level prompt, which we extend into a positive image-level prompt (PIP) and a negative image-level prompt (NIP). The edge image for PIP is the original edge image incorporating the depiction of a bounding box, while the sketch image for NIP is purely white. These two conditional inputs, denoted as $I^{'}_{e}$ and $\varnothing$, respectively, serve as the basis for the contrastive image-level prompt. They are then incorporated into the denoising process through the following equation:
\begin{equation}
\begin{aligned}
    &z_{t-1}=\widetilde{\epsilon}(z_t, I_{e}, I_{prompt})  \\
    &=\epsilon(z_t, \varnothing, \varnothing) +s_{cfg}(\epsilon(z_t, \varnothing, I_{prompt})-\epsilon(z_t, \varnothing, \varnothing)) \\
    &+s_{neg}(\epsilon^{'}(z_t, I_{e}, I_{prompt})-\epsilon(z_t, \varnothing, I_{prompt})), \\
    &\epsilon^{'}(z_t, I_{e}, I_{prompt}) = \epsilon(z_t, I_{e}, I_{prompt}) \\
    &+s_{pos}(\epsilon(z_t, I^{'}_{e}, I_{prompt})-\epsilon(z_t, I_{e}, I_{prompt})), \\
\end{aligned}
\end{equation}
where $s_{cfg}$ and $s_{neg}$ are used to finely adjust the respective effects of the PIP item and NIP item on the predictions, which will be discussed in our ablation study (see Fig. \ref{fig:IP_ablation}). PIP provides a subtle hint to the network, compelling it to include objects suitable for placing scene text in the generated image. On the other hand, NIP is used to control the clarity and visibility of the scene text. Through this contrastive image-level prompt, we provide the model with both a negative direction and a positive direction which enables the model to generate clear and precise scene text while maintaining a rational background.

\section{Experiments}
\subsection{Implementation Details}
\subsubsection{Experimental Settings}
Our model is built with Diffusers. The pre-trained models are ``runwayml/stable-diffusion-v1-5" and ``lllyasviel/sd-controlnet-canny".
While predicting, the size of the output images is $512\times512$. We use one A100 GPU for inference. The localized attention constraint is applied in both the U-Net branch and the control branch. The $\lambda$ in the localized attention constraint is 6.0. The $s_{cfg}$, $s_{neg}$ and $s_{cfg}$ are respectively 7.5, 2.0 and 0.1. The wordlist for localized attention constraint includes ``sign", ``billboard", ``label", ``promotions", ``notice", ``marquee", ``board", ``blackboard", ``slogan", ``whiteboard", and ``logo". 

\subsubsection{Evaluation}
 Due to the lack of publicly available multilingual benchmarks, we use multilingual vocabularies in the work of Zhang et al. \cite{zhang2021scene} and Xie et al. \cite{xie2023weakly} as the input texts and generate corresponding input prompts using chatGPT \cite{ouyang2022training}. We select five languages and filter out words with fewer than five characters. From the remaining set, we randomly choose 3000 words for each language. Ultimately, we generate 15,000 multilingual images for evaluation for each comparative method. We conduct both quantitative and qualitative comparative experiments. In the quantitative comparison, we utilize three metrics: CLIP Score \cite{hessel2021clipscore,huang2021unifying,radford2021learning,clipscore}, accuracy, and normalized edit distance \cite{shi2017icdar2017}. To ensure equitable capabilities across all languages for OCR tools, we use a multilingual OCR, namely easy-OCR \cite{easyocr}.
\subsection{Comparison with Existing Methods}
In this subsection, we compare our method with existing open-source methods capable of scene text generation, \textit{i.e.,} Stable Diffusion \cite{rombach2022high}, DeepFloyd \cite{deepfloyd}, TextDiffuser \cite{chen2023textdiffuser} and ControlNet \cite{zhang2023adding}. DeepFloyd uses two super-resolution modules to generate higher resolution $1024\times1024$ images compared with $512\times512$ images generated by other methods. We employ the template-to-image mode of the TextDiffuser method and utilize our sketch image as the template image. 

\subsubsection{Quantitative Comparison}
In the quantitative comparison, we selected the following three metrics: (1) \textbf{CLIPScore} is used to measure the similarity between the generated images and the input prompts. (2) \textbf{Accuracy evaluation} employs OCR tools to detect and calculate the recognition accuracy to assess whether the scene text in the generated images matches the input text. (3) \textbf{Normalized edit distance} is used to compare the similarity between the scene text in the generated images and the input text. We demonstrate the quantitative results compared with existing methods in Table \ref{table:quantitative}. As shown in Table \ref{table:quantitative}, Although training-free, our method still achieves a competitive CLIP score and significantly enhances the recognition accuracy of generated images. For each specific language, our method demonstrates an average improvement in accuracy of 25\% compared to the existing method.

\begin{figure}[!t]
    \centering
    \includegraphics[width=0.98\linewidth]{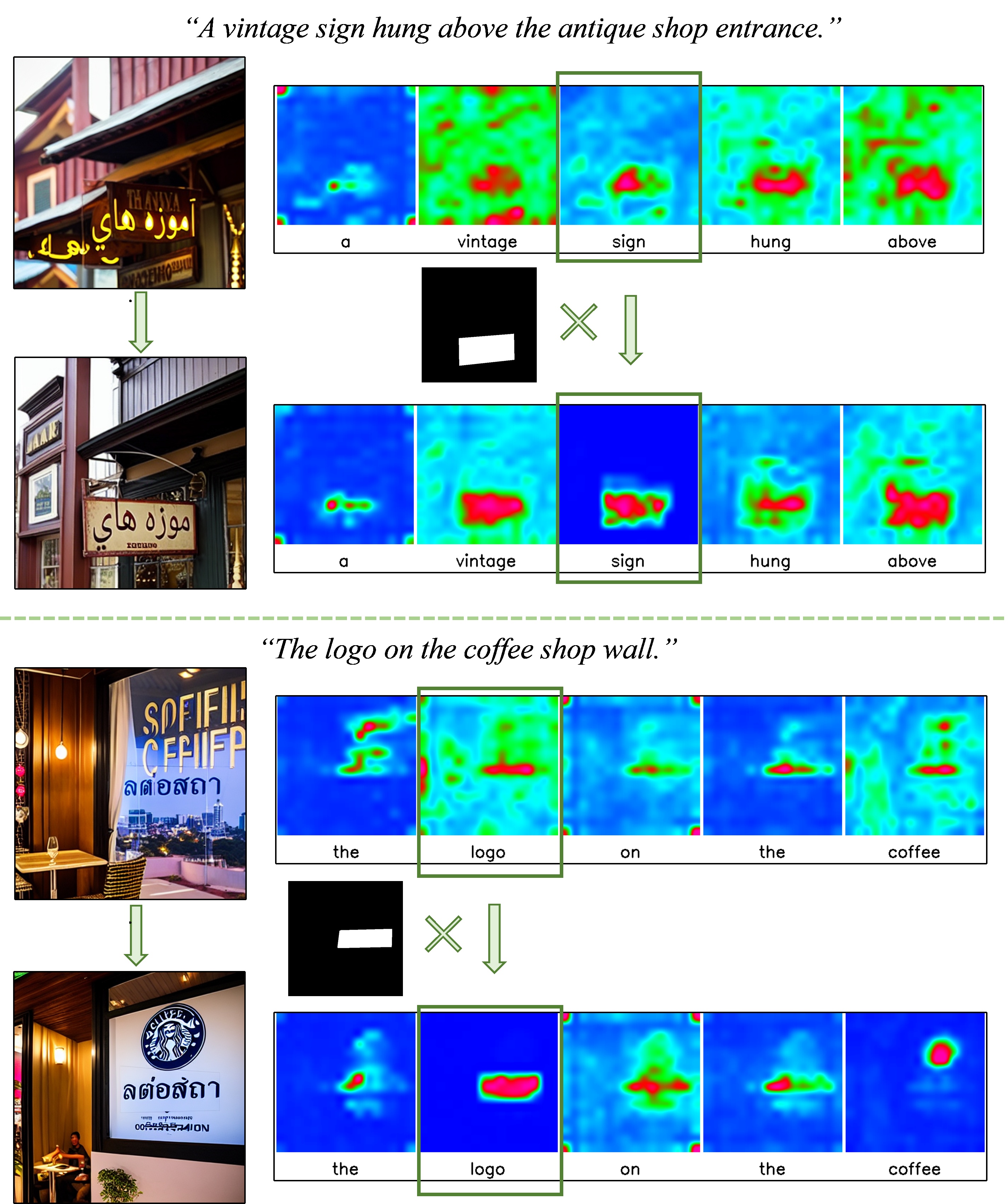}
    \caption{Visualization of ablation experiments on the localized attention constraint method. The heatmaps illustrate the average cross-attention map corresponding to different tokens across all diffusion steps.}
    \label{fig:attn_ablation}
\end{figure}

\subsubsection{Qualitative comparison}
Fig. \ref{fig:big_compare} and Fig. \ref{fig:small_compare} show the comparison between our method and existing methods in generating scene text images for majority and minority languages, respectively. From Fig. \ref{fig:big_compare}, it can be observed that for English, which has a significant presence in the training dataset of existing methods, the generated images possess a certain level of recognizability. However, Stable Diffusion and DeepFloyd may exhibit instances of generating multiple or missing characters. TextDiffuser, with the sketch image as an input template, addresses the issue of multiple and missing characters. Nevertheless, due to insufficient strictness in control, TextDiffuser still encounters problems with erroneous character generation. Despite utilizing edge images for strict control, ControlNet still results in the generation of scene text appearing in unreasonable positions or having additional strokes. In contrast, our method can generate clear, precise, and reasonably positioned scene text. For the languages with a smaller presence in the training dataset (Chinese, Arabic, Thai, Russian), Stable Diffusion, DeepFloyd, and TextDiffuser fail to generate recognizable scene text. TextDiffuser may generate some English letters instead of similar characters from other languages. ControlNet still encounters issues of generating text in unreasonable positions, and when dealing with characters resembling special patterns, such as Arabic characters, ControlNet merges the text with the background, rendering the generated text unidentifiable. Our method, on the other hand, successfully generates scene text images for all languages.

\subsection{Ablation Study}
To validate the effectiveness of the proposed localized attention constraint and contrastive image-level prompt, we conduct the ablation study. Table \ref{table:ablation} presents the quantitative analysis of the ablation experiments. As demonstrated in Table \ref{table:ablation}, it is evident that the full model achieves the best performance in both the CLIP score and the accuracy of generated characters. In addition, we also conduct qualitative analysis for the ablation study, and the results are presented in Fig. \ref{fig:IP_ablation} and \ref{fig:attn_ablation}. The seed is fixed at 2345 to generate visualized results. In Fig. \ref{fig:IP_ablation}, we discuss the impact of different parameters for PIP and NIP (i.e., $s_{pos}$ and $s_{neg}$) on the generated images. From Fig. \ref{fig:IP_ablation}, it can be observed that as $s_{pos}$ increases, the sign in the background becomes more prominent, but excessively high $s_{pos}$ can cause the sign to appear too pronounced and flat. On the other hand, as $s_{neg}$ increases, the scene text in the foreground becomes clearer, but excessively high $s_{neg}$ can result in scene text floating in unreasonable positions. Fig. \ref{fig:attn_ablation} showcases the results with and without our localized attention constraint. It can be observed that when we constrain the cross-attention map corresponding to the ``sign" and ``logo" to the scene text region, the generated images appear more reasonable and realistic.


\begin{table}[!t]
\center
\resizebox{\linewidth}{11.5mm}{
\begin{tabular}{cccc}
\toprule
Method                          & CLIP            & Accuracy       & Edit accuracy  \\ \midrule
W/o constraint            & 0.8065          & 31.42          & 74.30          \\
W/o PIP & 0.7935          & 27.68          & 70.92         \\
W/o NIP & 0.7718          & 10.39          & 50.95          \\
\textbf{Full model}             & \textbf{0.8108} & \textbf{35.48} & \textbf{77.22} \\ \bottomrule
\end{tabular}
}
\caption{Quantitative ablation studies on localized attention constraint and image level prompt. ``W/o constraint" denotes the exclusion of the localized attention constraint method, ``W/o NIP" denotes the exclusion of the negative image-level prompt, and ``W/o PIP" denotes the exclusion of the positive image-level prompt.
The results indicate that our full model achieves the best generation results.}
\label{table:ablation}
\end{table}
\section{Discussion and Conclusion}
Currently, the bounding box of the text region is obtained either through user specification or random generation, and the tokens in the prompt that require localized attention constraint are determined by manually given wordlists. In future work, it is possible to integrate these two parts with GPT4 API for a more rational selection of bounding boxes and wordlists. However, our model still faces challenges in generating small-scale scene text and achieving precise text color control. Moreover, the generated scene text still occasionally includes unintended textual elements.

In this paper, we introduce a training-free framework, named Diff-Text. This framework is designed to apply to scene text generation in any language. Localized attention constraint method and contrastive image-level prompt are proposed to enhance the precision, clarity, and coherence of generated scene text images.
\bigskip

\section{Acknowledgement}
This work was jointly supported by the National Natural Science Foundation of China under Grant No. 62102150, No. 62176091, the National Key Research and Development Program of China under Grant No. 2020AAA0107903.
\bibliography{aaai24}
\clearpage
\appendix
\section{Appendix}
We provide more details of the proposed method and additional experimental results to help better understand our paper. In summary, this appendix includes the following contents:
\begin{itemize}
    \item Contrastive image-level prompts details.
    \item Evaluation metrics.
    \item Details of compared methods.
    \item Limitations of our model.
    \item Involving of GPT4.
    \item More results of our proposed method.
\end{itemize}
\begin{table}[th]
\centering
\begin{tabular}{ccc}
\toprule
Language & \multicolumn{1}{l}{Accuracy} & \multicolumn{1}{l}{Edit\_accuracy} \\ \midrule
English  & 92.80                         & 98.52                              \\
Chinese  & 60.85                        & 82.47                              \\
Arabic   & 60.67                         & 88.95                              \\
Thai     & 72.05                           & 93.99                              \\
Russian  & 53.67                         & 87.07                              \\ \bottomrule
\end{tabular}
\caption{The evaluation of OCR tool.}
\label{table:ocr}
\end{table}
\subsection{Contrastive Image-level Prompts Details}
As discussed before, we consider the edge image $I_{e}$ as the image-level prompt. The combination of the image-level prompt input $I_{e}$ and text prompt input $I_{prompt}$ constitutes classifier-free guidance with two conditions. Specifically, our model can be formulated as follows:
\begin{equation}
    z_{t-1} = \epsilon(z_t, I_{e}, I_{prompt}),
\end{equation}
By Bayes' theorem, we can derive:
\begin{equation}
\begin{aligned}
    &P(z_t|I_{e},I_{prompt}) \\ 
    &=\frac{P(I_{e}|z_t,I_{prompt})P(I_{prompt}|z_t)P(z_t)}{P(I_{e},I_{prompt})} \\
\end{aligned}
\end{equation}
By taking the logarithm and differentiating both sides of the equation simultaneously, we can obtain the result:
\begin{equation}
\begin{aligned}
    &\nabla_{z_{t}}log(P(z_t|I_{e},I_{prompt})) \\
    &=\nabla_{z_{t}}log(P(z_t)) \\
    &+\nabla_{z_{t}}log(P(I_{prompt}|z_t)) \\
    &+\nabla_{z_{t}}log(P(I_{e}|z_t,I_{prompt})), \\
\end{aligned}
\end{equation}
which can also be represented by:
\begin{equation}
\begin{aligned}
    &z_{t-1}=\widetilde{\epsilon}(z_t, I_{e}, I_{prompt})  \\
    &=\epsilon(z_t, \varnothing, \varnothing) +s_{cfg}(\epsilon(z_t, \varnothing, I_{prompt})-\epsilon(z_t, \varnothing, \varnothing)) \\
    &+s_{neg}(\epsilon^{'}(z_t, I_{e}, I_{prompt})-\epsilon(z_t, \varnothing, I_{prompt})), \\
\end{aligned}
\end{equation}
In our implementation, we achieve $\epsilon(z_t, \varnothing, I_{prompt})$ by using a purely white image as the image-level prompt, which we denote as the negative image-level prompt (NIP). Furthermore, an additional bounding box with a width of 1 pixel is drawn in the edge image and is considered as a positive image-level prompt (PIP). The denoised result conditioned on the PIP is represented as $\epsilon(z_t, I^{'}{e}, I{prompt})$. Similarly, the denoised result conditioned on the original image-level prompt is represented as $\epsilon(z_t, I_{e}, I_{prompt})$. These two conditions are combined using the following equation:
\begin{equation}
\begin{aligned}
    &\epsilon^{'}(z_t, I_{e}, I_{prompt}) = \epsilon(z_t, I_{e}, I_{prompt}) \\
    &+s_{pos}(\epsilon(z_t, I^{'}_{e}, I_{prompt})-\epsilon(z_t, I_{e}, I_{prompt})), \\
\end{aligned}
\end{equation}
Through this approach, we provide the model with both a negative direction, represented by the purely white image, and a positive direction, represented by the presence of signs in the text region.

\subsection{Evaluation Metrics}
To evaluate the performance of the proposed Diff-Text, we use three metrics, including CLIP score, accuracy, and normalized edit distance. The details are as follows:
\begin{itemize}
    \item \textbf{CLIP score} is employed to measure the similarity between the generated images and the input prompts. We used the off-the-shelf calculation code in \cite{clipscore}. It is noteworthy that the input prompts utilized within our approach do not encompass textual compositions, which is different from Stable Diffusion \cite{rombach2022high} and DeepFloyd \cite{deepfloyd}. In the computation of the CLIP score, we calculate the similarity between the generated scene text image and the augmented input prompt, which incorporates textual content suffixed with "that reads `xxx'." For instance, we substitute the phrase "A sign on the street" with "A sign that reads `xxx' on the street." For the compared method TextDiffuser and ControlNet whose captions are the same as ours, we apply the same strategy.
    \item \textbf{Accuracy evaluation} employs OCR tools to detect and calculate the recognition accuracy to assess whether the scene text in the generated images matches the input text. To ensure equitable capabilities across all languages for OCR tools, we use a multilingual OCR, namely easyOCR \cite{easyocr}. 
    \item \textbf{Normalized edit distance} is utilized for assessing the similarity between the text recognized by the OCR tool and the ground truth. This metric quantifies similarity by enumerating the minimal set of operations needed to transform one string into another. Specifically, the normalized edit distance is computed as:
    \begin{equation}
        Norm=1-\frac{D(s_{i},\Tilde{s_{i}})}{MaxLen(s_{i},\Tilde{s_{i}})},
    \end{equation}
    where $D(\cdot)$ indicates the Levenshtein Distance. $\Tilde{s_{i}}$ and $s_{i}$ indicate the predicted scene text in string and the corresponding ground truths. $MaxLen$ represents the maximum length between the strings $\Tilde{s_{i}}$ and $s_{i}$.
\end{itemize}
Due to the inherent constraints in the detection and recognition capability of the OCR tool, the presented accuracy and edit distance data within the paper are easy to be understated compared to actual values. To ascertain the ability of the employed OCR tools, we conduct assessments using sketch images, and the results of these tests are delineated in Table \ref{table:ocr}.

\begin{figure}[h]
    \centering
    \includegraphics[width=1.0\linewidth]{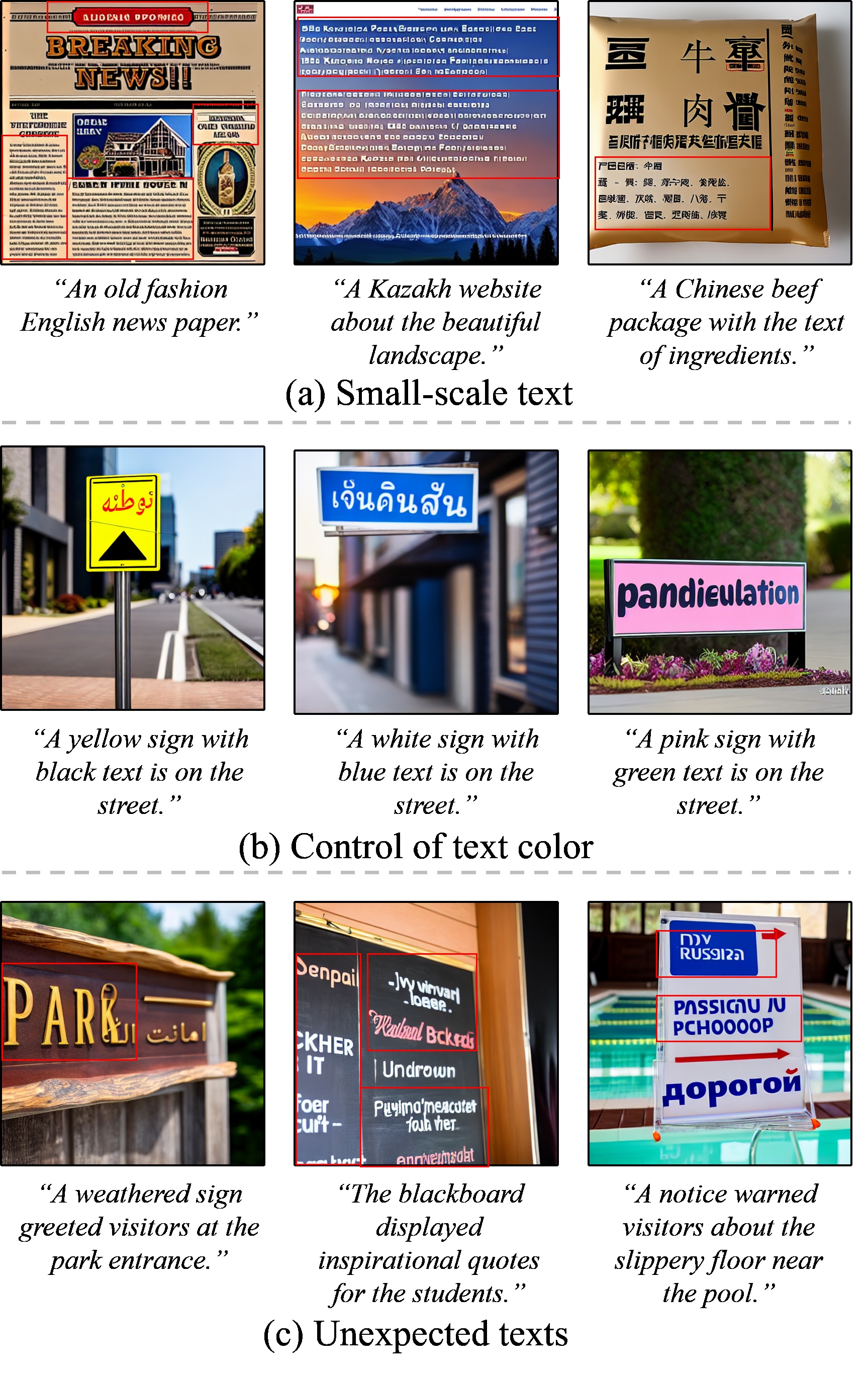}
    \caption{Limitations of our method. The region enclosed within the red rectangular border represents the inaccurately generated segment.}
    \label{fig:limitation}
\end{figure}
\subsection{Details of Compared Methods}
We compare the proposed model with Stable Diffusion, ControlNet, DeepFloyd, and TextDiffuser. The details of compared methods are as follows:
\begin{itemize}
\item \textbf{Stable Diffusion} \cite{rombach2022high} is an open-sourced model that projects the image into latent space with VAE and applies the diffusion process to generate feature maps in the latent level. This model uses a CLIP \cite{hessel2021clipscore,huang2021unifying,radford2021learning} text encoder for the acquisition of user prompt embeddings. The publicly available pre-trained model identified as ``runwayml/stable-diffusion-v1-5" is used. The number of sampling steps is 50, and the scale of classifier-free guidance is 7.5. The input prompts contain textual content.
\item \textbf{ControlNet} \cite{zhang2023adding} is a model used to provide more control for Stable Diffusion. The extra condition is introduced by zero convolution. We use the pre-trained model identified as ``lllyasviel/sd-controlnet-canny". The control scale is 1.0, the number of sampling steps is 50, and the scale of classifier-free guidance is 7.5. The input prompts are the same as ours and contain no textual content.
\item \textbf{DeepFloyd} \cite{deepfloyd} employs three cascaded diffusion modules to generate images of progressively augmented resolution: 64x64, 256x256, and 1024x1024. All stage modules use T5 Transformer as the text encoders. We use default models and parameters for inference, where the three pre-trained cascaded models are ``DeepFloyd/IF-I-XL-v1.0”, ``DeepFloyd/IF-II-L-v1.0”, and ``stabilityai/stable-diffusion-x4-upscaler”. The input prompts contain textual content.
\item \textbf{Text Diffuser} \cite{chen2023textdiffuser} is a two-stage model containing a transformer to generate character-level masks and a latent diffusion model controlled by the masks. We use the official implementation on GitHub and use the ``text-to-image-with-template" mode. The input prompts contain no textual content.
\end{itemize}
\subsection{Limitations of Our Model}
It is observed that our model still struggles to generate small-scale scene text and achieve precise text color control. Moreover, the generated scene text still occasionally includes unintended textual elements. We display the failure cases generated by our model in Fig. \ref{fig:limitation}. During the generation of small-scale text, our model produces strokes that lack clarity, resulting in the text becoming unrecognizable. Additionally, our model cannot effectively control the color of the scene text. Moreover, we have observed instances where the generated scene text includes unexpected English text.
\begin{figure}
    \centering
    \includegraphics[width=1.0\linewidth]{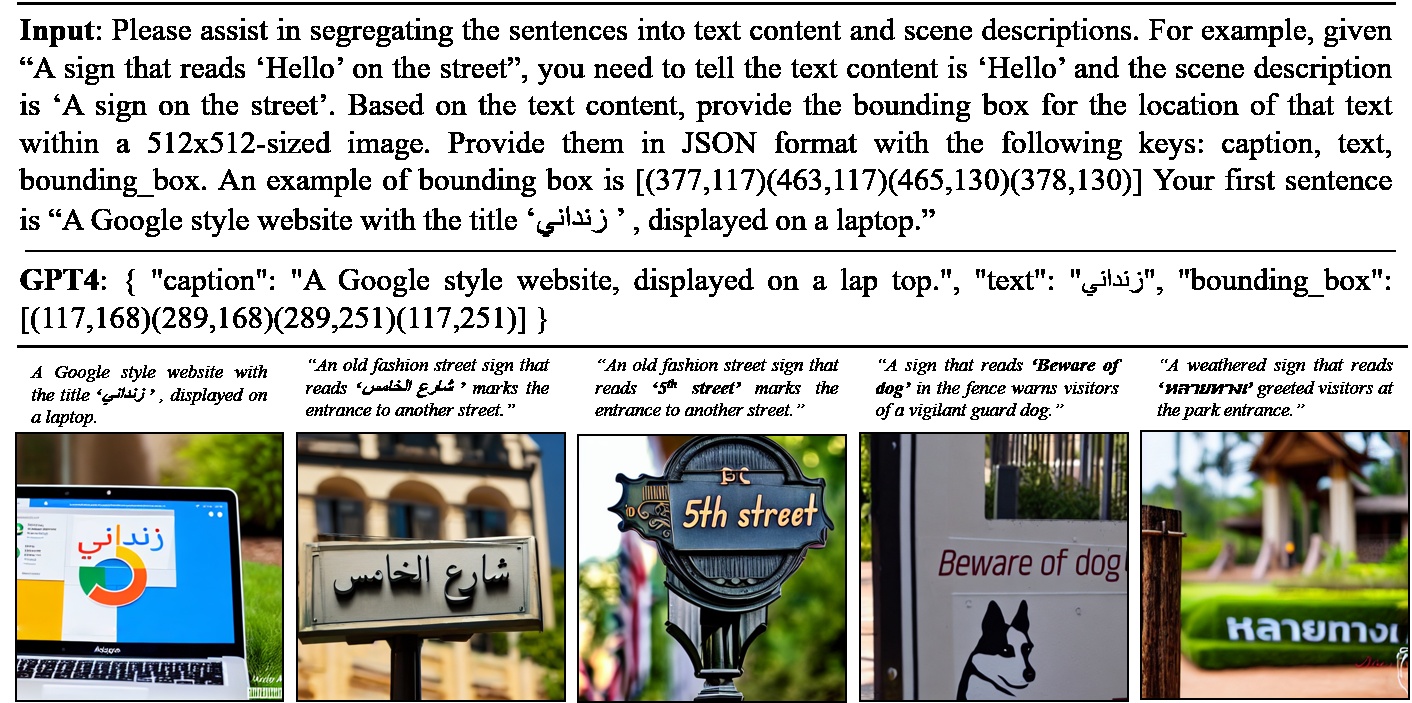}
    \caption{Results involving GPT4.}
    \label{fig:gpt4}
\end{figure}
\subsection{Involing of GPT4}
As we mentioned in our discussion, the tokens in the prompt that require localized attention constraints are determined by a pre-defined wordlist. We offer an alternative way by using GPT4, which can be seen in Fig. \ref{fig:gpt4}. GPT-4 automatically identifies tokens suitable as carriers for scene text, enhancing the flexibility of generation.

\begin{figure}[!th]
    \centering
    \includegraphics[width=1.0\linewidth]{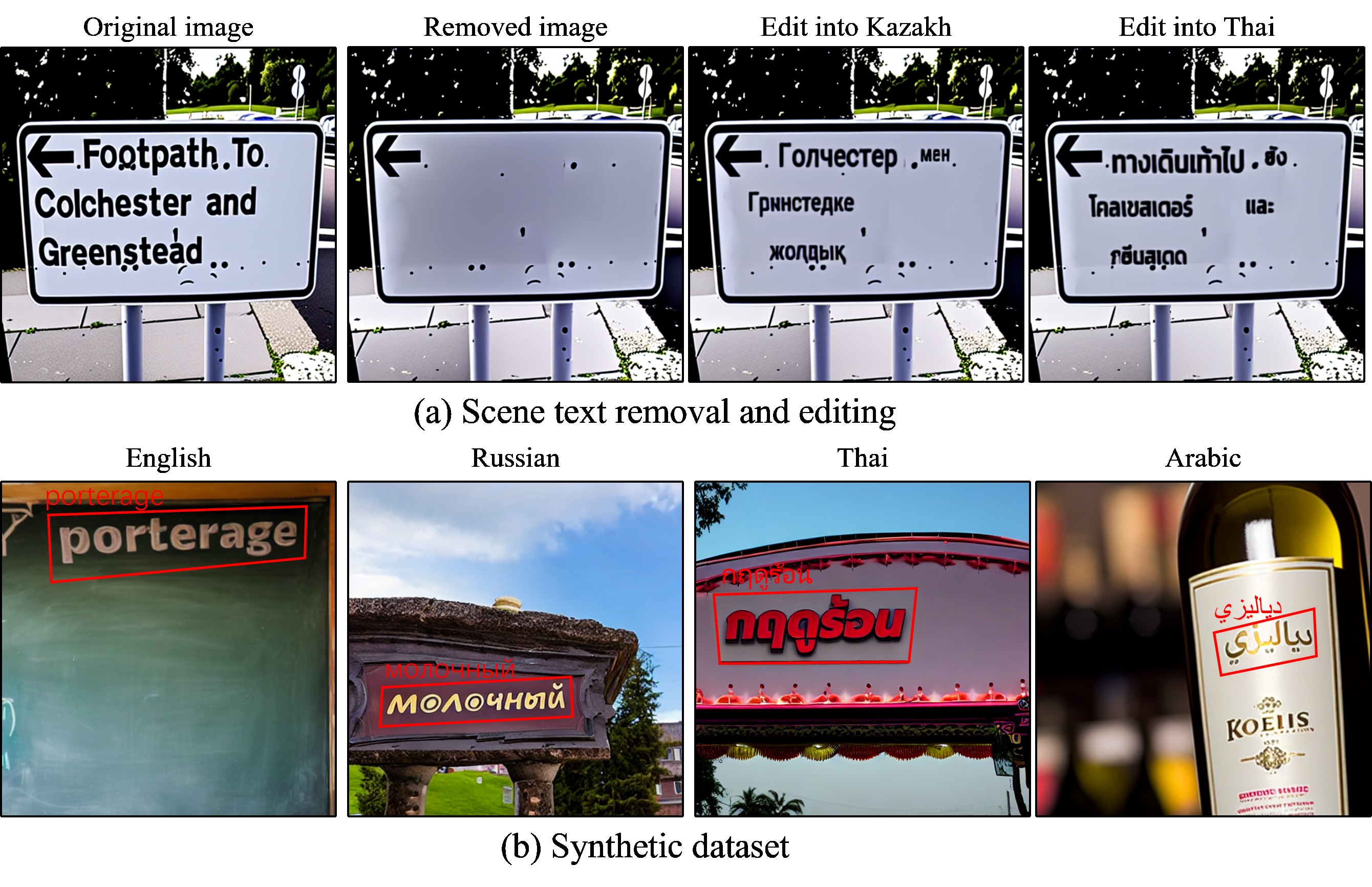}
    \caption{The Applications of Our Model. Our model can be employed for (a) scene text removal and scene text editing, and (b) synthesizing multilingual scene text datasets.}
    \label{fig:discussion}
\end{figure}
\begin{figure}[!h]
    \centering
    \includegraphics[width=1.0\linewidth]{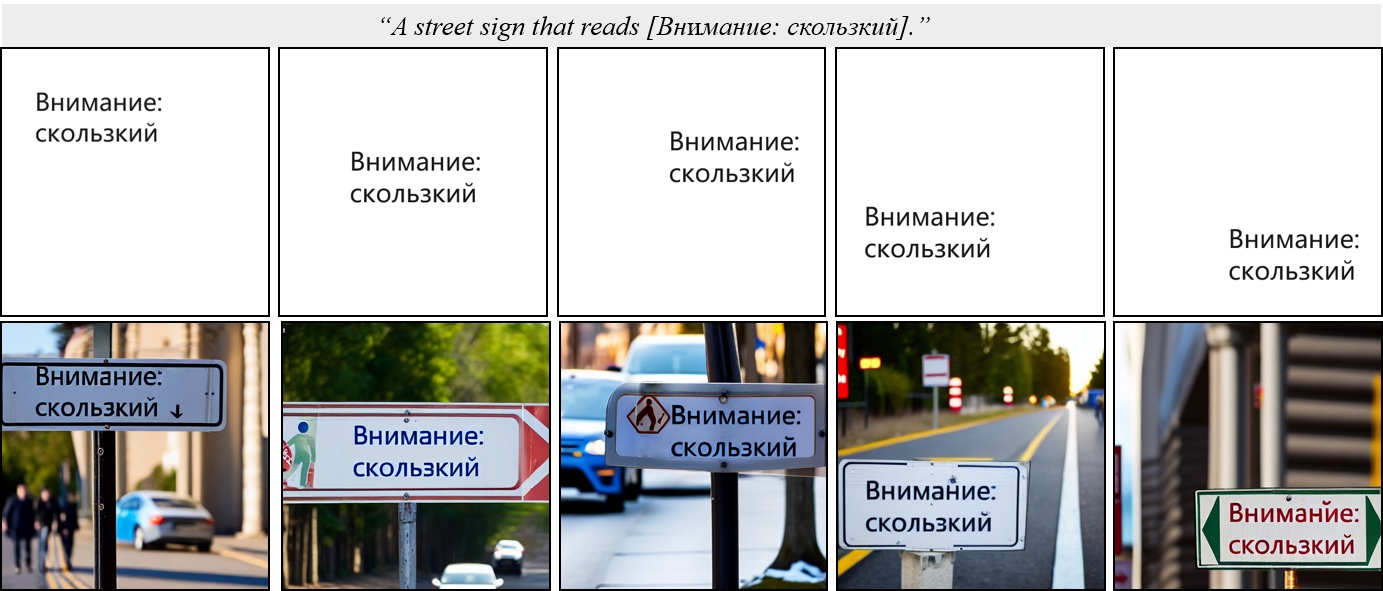}
    \caption{Results of various text positions. The top row shows the sketch image, and the bottom row displays the corresponding output image.}
    \label{fig:bbx}
\end{figure}
\subsection{Applications}
As shown in Fig. \ref{fig:discussion}, our method is capable of scene text removal, scene text editing, and synthetic dataset generation. Moreover, as illustrated in Figure \ref{fig:bbx}, the images generated by our model can accommodate various text positions.

\subsection{More results of proposed method}
\begin{figure*}[th]
    \centering
    \includegraphics[width=\textwidth]{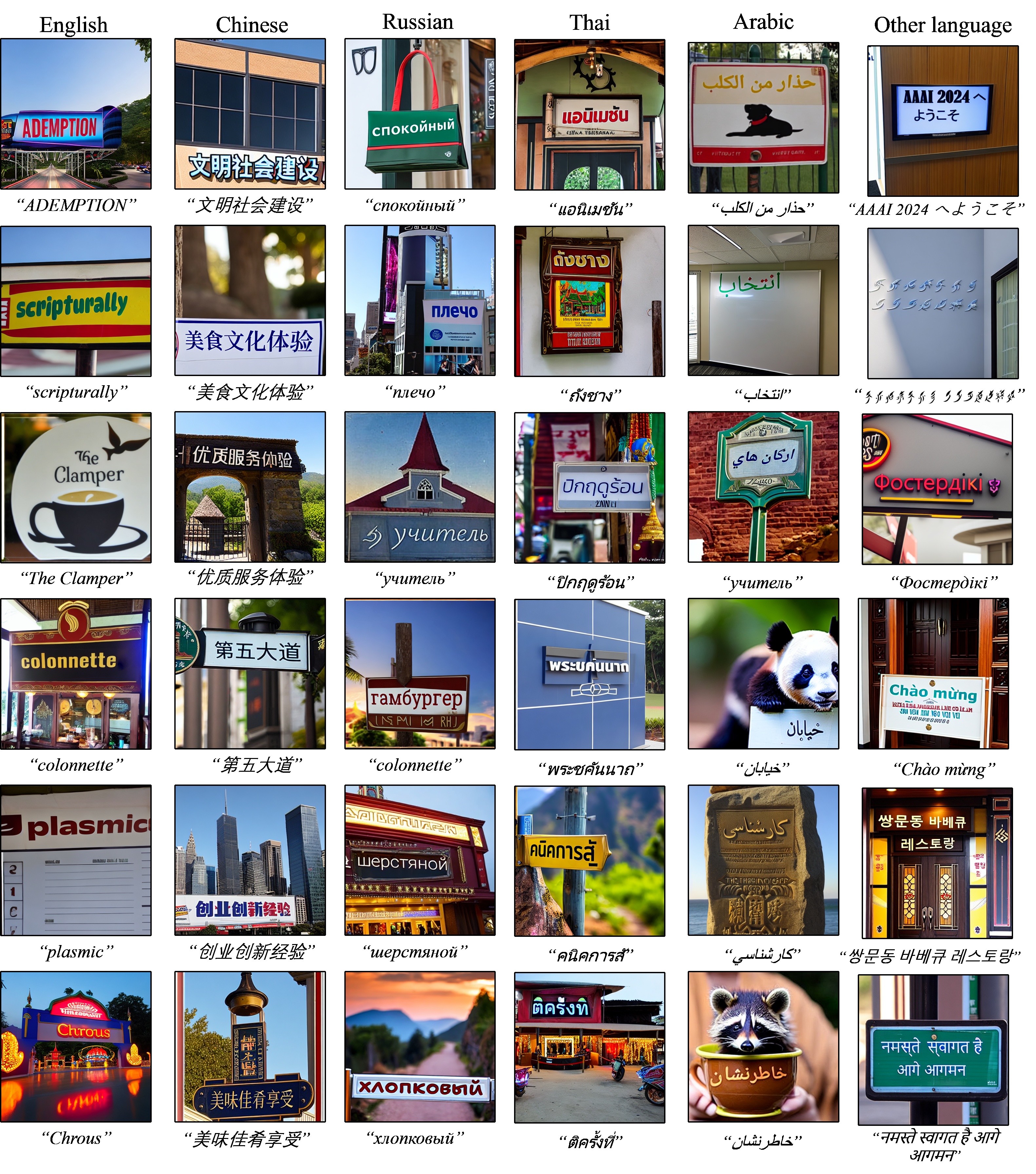}
    \caption{More results of the proposed method.}
    \label{fig:more}
\end{figure*}

In this subsection, we display more results of the proposed method. As shown in Fig. \ref{fig:more}, we generate 11 languages using our proposed model, including English, Chinese, Russian, Thai, Arabic, Japanese, Nvshu, Kazakh, Vietnamese, Korean, and Hindi. 

\end{document}